\pdfoutput=1

\documentclass[11pt]{article}

\usepackage{acl}

\usepackage{times}
\usepackage{latexsym}
\usepackage{amssymb, amsmath, amsfonts}
\usepackage{multirow}
\usepackage{booktabs}
\usepackage{comment}

\usepackage[T1]{fontenc}

\usepackage[utf8]{inputenc}

\usepackage{microtype}

\usepackage{listings}

\lstnewenvironment{lispcode}{
  \lstset{language=Lisp}
}{}

\usepackage{xcolor} %
\definecolor{LightGray}{gray}{0.9}

\definecolor{backcolour}{RGB}{245,248,250}
\definecolor{emph}{RGB}{166,88,53}
\definecolor{nightblue}{RGB}{9,49,105}
\definecolor{keywords}{RGB}{207,33,46}
\definecolor{lightpurple}{RGB}{130,81,223}

\lstdefinestyle{prompt}{
    backgroundcolor=\color{backcolour},
    commentstyle=\color{codegreen},
    keywordstyle=\color{keywords},
    stringstyle=\color{nightblue},
    basicstyle=\fontsize{7}{8}\ttfamily,
    breakatwhitespace=true,
    breaklines=true,
    captionpos=b,
    keepspaces=true,
    numberstyle=\tiny\color{codegray},
    numbersep=2pt,
    showspaces=false,
    showstringspaces=false,
    showtabs=false,
    tabsize=2,
    linewidth=0.98\columnwidth,
    frame=tb,
    xrightmargin=0pt,
    xleftmargin=0.23cm,
    numbers=none,
    aboveskip=0.4cm,
    belowskip=0.4cm,
}

\renewcommand{\vec}[1]{\mathbf{\boldsymbol{#1}}}

\newcommand{\hvec}{\vec{h}}

\newcommand{\rvec}{\vec{r}}

\newcommand{\uvec}{\vec{u}}

\newcommand{\xvec}{\vec{x}}
\newcommand{\yvec}{\vec{y}}

\usepackage[disable]{todonotes}

\makeatletter
\newcommand*\iftodonotes{\if@todonotes@disabled\expandafter\@secondoftwo\else\expandafter\@firstoftwo\fi}  %
\makeatother

\iftodonotes{
    \paperwidth=\dimexpr\paperwidth + 12cm\relax
    \oddsidemargin=\dimexpr\oddsidemargin + 6cm\relax
    \evensidemargin=\dimexpr\evensidemargin + 6cm\relax
    \marginparwidth=\dimexpr\marginparwidth + 4cm\relax
    \marginparsep=\dimexpr\marginparsep + 2cm\relax
}{}

\newcommand{\emldisplay}[2]{\texttt{\href{mailto:#1}{#2}}}
\newcommand{\eml}[1]{\emldisplay{#1}{#1}}
\newcommand{\eg}{\emph{e.g.,}\ }
\newcommand{\ie}{\emph{i.e.,}\ }

\newcommand{\datasetfullname}{\textsl{SMCalFlow-EventQueries}}
\newcommand{\dataset}{\textsl{SMCalFlow-EQ}}

\title{Few-Shot Adaptation for Parsing Contextual Utterances with LLMs}

\author{
  Kevin Lin $^\dag$\thanks{\; Work done at Microsoft Semantic Machines.}
  \quad
  Patrick Xia $^\ddag$
  \quad
  Hao Fang $^\ddag$
  \\
  $^\dag$ UC Berkeley
  \quad
  $^\ddag$ Microsoft Semantic Machines 
  \\
  \eml{k-lin@berkeley.edu}
  \quad
  \tt{\{\emldisplay{patrickxia@microsoft.com}{patrickxia}, \emldisplay{hao.fang@microsoft.com}{hao.fang}\}@microsoft.com}
}

\begin{document}

\maketitle

\begin{abstract}

We evaluate the ability of semantic parsers based on large language models (LLMs) to handle contextual utterances.
In real-world settings, there typically exists only a limited number of annotated contextual utterances due to annotation cost, resulting in an imbalance compared to non-contextual utterances. 
Therefore, parsers must adapt to contextual utterances with a few training examples.
We examine four major paradigms for doing so in conversational semantic parsing \ie Parse-with-Utterance-History, Parse-with-Reference-Program, Parse-then-Resolve, and Rewrite-then-Parse.
To facilitate such cross-paradigm comparisons, we construct \datasetfullname{}, a subset of contextual examples from SMCalFlow with additional annotations. 
Experiments with in-context learning and fine-tuning suggest that Rewrite-then-Parse is the most promising paradigm when holistically considering parsing accuracy, annotation cost, and error types.

\end{abstract}
\section{Introduction}

A key challenge in conversational semantic parsing (CSP) is 
handling {\it contextual} utterances (\ie utterances that can only be understood with its context) by mapping them to {\it non-contextual} programs that can be fulfilled by an executor without relying on the dialogue state.
Many approaches have been proposed, \eg 
directly mapping the contextual utterance with utterance history 
to a non-contextual program \cite{suhr-etal-2018-learning},
or mapping to an intermediate contextual program which is then resolved
(usually in a deterministic manner) to a non-contextual program \cite{SMDataflow2020,cheng-etal-2020-conversational}.
In these prior works, there is often an assumption of having a substantial corpus of annotated data 
encompassing both non-contextual utterances and contextual utterances for training a parser.
However, in practice, it is more expensive to collect and annotate contextual utterances 
compared to non-contextual utterances, due to the dependency on the conversation history.
Furthermore, annotating non-contextual utterances usually precedes annotating contextual utterances.
To reflect such real-world settings, we study few-shot adaptation for parsing contextual utterances, 
where we first build a parser using a large number of annotated non-contextual utterances,
and then adapt it for parsing contextual utterances using a few (or even zero) annotated contextual utterances.

Recent work has shown that large language models (LLMs) are capable of 
semantic parsing using a few examples \cite{shin-etal-2021-constrained,shin-van-durme-2022-shot}.
Hence, in this work, we conduct a focused study on few-shot adaptation using LLMs for CSP.
Specifically, we consider four major paradigms:
Parse-with-Utterance-History, Parse-with-Reference-Program, Parse-then-Resolve, and Rewrite-then-Parse.
One challenge of carrying out a comparative study on these paradigms is the lack of annotated data,
since existing CSP datasets such as SMCalFlow \cite{SMDataflow2020} and CoSQL \cite{yu-etal-2019-cosql}
are often annotated based on a single paradigm.
Therefore, we construct a new dataset, \dataset{}, derived from a subset of SMCalFlow dialogues
with annotations for all four paradigms.

\begin{figure*}[!t]
    \centering
    \includegraphics[width=.98\textwidth,clip,trim=0in 2.8in 0in 0in]{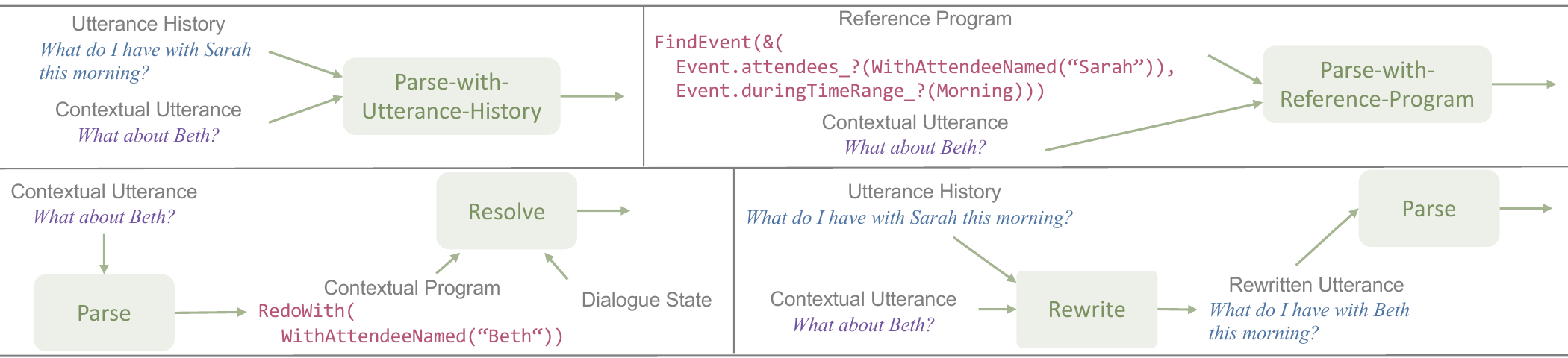}
    \caption{Four canonical paradigms of conversational semantic parsing for contextual utterances.}
    \label{fig:paradigms}
\end{figure*}

Our experiments consider both in-context learning (ICL) using GPT-3.5 %
and fine-tuning (FT) using T5-base 220M \cite{raffel2020exploring}
for building and adapting parsers. ICL typically has lower accuracy compared to FT, although the two are not strictly comparable as they use different models. The only exception is Parse-with-Reference-Program, suggesting that GPT-3.5 is effective at 
editing programs using natural language. Overall, we find Rewrite-then-Parse to be the most promising approach, as it achieves similar accuracy to other paradigms in both ICL and FT experiments,
while requiring only a few annotated examples for to develop a query rewriter and 
no additional program annotations.
We release code and data to facilitate future work on parsing contextual utterances.\footnote{%
\url{https://github.com/microsoft/few_shot_adaptation_for_parsing_contextual_utterances_with_llms}}

\section{Background: LLM-Based Parsing}
\label{sec:llm_parsing}

Following \citet{shin-etal-2021-constrained} and \citet{roy-etal-2022-benchclamp}, 
we formulate parsing as a constrained decoding problem, 
where an LLM is used to predict the next token 
and a context-free grammar (CFG) is used to validate the predicted token.
A program is represented as a sequence of S-expression tokens $y_1 y_2 \ldots y_L$.
The space of all valid S-expressions is governed by a CFG denoted by $\mathcal{G}$,
which can be automatically derived from function definitions and types used in the domain (see Appendix~\ref{appendix:cfg}).

To generate the program for a user utterance,
we first feed the LLM with the user utterance and necessary context information as a sequence of tokens.
Then the S-expression of the program is generated incrementally.
At each decoding step $l$, we only keep the partial prefix sequence $y_1 y_2 \ldots y_l$ if it is allowed by $\mathcal{G}$.
This validation can be efficiently performed via Earley's parsing algorithm \cite{earley-1970-efficient}
using the parsing state of the partial sequence $y_1 y_2 \ldots y_{l-1}$.

In this paper, we consider both ICL and FT for constructing LLM-based parsers.
For ICL, we prompt the pre-trained LLM with 
$K_{\textrm{ICL}}$ demonstration examples retrieved via BM25 \cite{bm25-1994-simple,bm25-2009-probablistic}, following \citet{rubin-etal-2022-learning} and \citet{roy-etal-2022-benchclamp}.
For FT, we continue training the LLM on $K_{\textrm{FT}}$ demonstration examples, producing a new model to be used during constrained decoding.

\section{Few-Shot Adaptation}
\label{sec:few-shot}

In this paper, we assume there are a large number ($M$) of annotated non-contextual utterances,
$\mathcal{D} = \{ (\xvec^{(1)}, \yvec^{(1)}), \ldots, (\xvec^{(M)}, \yvec^{(M)}) \}$,
where $\xvec^{(i)}$ denotes the $i$-th non-contextual utterance in the dataset,
$\yvec^{(i)}$ is the corresponding non-contextual program, and $M$ is the number of annotated examples.
These examples are used to derive a grammar $\mathcal{G}_1$ and build the parser $\mathcal{P}_1$ for non-contextual utterances via either ICL or FT.

For a contextual utterance $\uvec_t$ at the $t$-th turn of a dialogue, 
the goal is to obtain the non-contextual program $\yvec_t$
using the \textit{utterance history} $\hvec_t = [\uvec_{<t}]$, the corresponding programs $\yvec_{<t}$,
and/or other information recorded in the dialogue state.
\autoref{fig:paradigms} illustrates four canonical paradigms for parsing contextual utterances.
For each of these paradigms, we would like to obtain a new parser by adapting from the base parser $\mathcal{P}_1$ using $N$
demonstration examples, where $N \ll M$.

\subsection{Parsing Paradigms}

\noindent
{\bf Parse-with-Utterance-History}:
In this paradigm, the parser directly predicts $\yvec_t$ by conditioning on the contextual utterance $\uvec_t$ and its history $\hvec_t$.
This paradigm has been used in contextual semantic parsing \cite{zettlemoyer-collins-2009-learning,suhr-etal-2018-learning} and belief state tracking \cite{mrksic-etal-2017-neural}.

\noindent
{\bf Parse-with-Reference-Program}:
This paradigm assumes that the salient additional context to parse $\uvec_t$ is captured by a {\it reference program},
which is a non-contextual program to be revised and typically that from the preceding turn, $\yvec_{t-1}$. 
The parsing process can be viewed as editing the reference program based on the contextual utterance
which directly yields $\yvec_t$.
\citet{zhang-etal-2019-editing} employs a similar strategy by using a copy operation during parsing 
to copy tokens from the reference program for text-to-SQL.

\noindent
{\bf Parse-then-Resolve}:
This paradigm divides the task into two steps, leading to a modularized
system with a parser followed by a resolver.
$\uvec_t$ is first mapped to an intermediate program $\tilde{\yvec}_t$ which contains specialized contextual symbols. 
These contextual symbols (marking ellipsis or coreference) are resolved deterministically using the dialogue state determined from $\yvec_{<t}$, resulting in the final non-contextual prediction $\yvec_t$.
Several recent datasets for CSP have adopted this paradigm \cite{SMDataflow2020,cheng-etal-2020-conversational}.

\noindent
{\bf Rewrite-then-Parse}:
This paradigm modularizes the system using a rewriter followed by a parser.
The history $\hvec_t$ and contextual utterance $\uvec_t$ are first rewritten into a single non-contextual utterance $\uvec'_t$
Then, $\uvec'_t$ is parsed to $\yvec_t$ by a single-turn semantic parser.
This paradigm is closely related to incomplete utterance rewriting \cite{liu-etal-2020-incomplete}
and conversational query rewriting \cite{rastogi-etal-2019-scaling,yu-etal-2020-few-shot, chen2020pretraining, song2020two-stage,inoue-etal-2022-enhance,mao2023large}
though the parsing step is usually unnecessary or overlooked in these related studies.
Using this paradigm, the rewriter and the parser can be independently developed and maintained.

\subsection{Adaptation via ICL}

For ICL, we use GPT-3.5 and 
the following prompt template provided by 
\citet{shin-etal-2021-constrained} and \citet{roy-etal-2022-benchclamp},
where placeholders \{X1\}, \{X2\}, \ldots\ are demonstrations input,
\{Y1\}, \{Y2\}, \ldots\ are demonstrations output,
and \{X$^\prime$\} is the test input.
\begin{lstlisting}[breaklines=true]
Let's translate what a human user says into what
a computer might say.


Human: {X1}
Computer: {Y1}

Human: {X2}
Computer: {Y2}

...

Human: {X'}
Computer:
\end{lstlisting}

For Parse-with-Utterance-History, Parse-with-Reference-Program, and Parse-then-Resolve,
the input placeholders are respectively instantiated as
$\hvec\ \lvert\ \uvec$, $\rvec\ \lvert\ \uvec$, and $\uvec$, 
where the character $\lvert$ is used as the separator.
The output placeholders are all instantiated by non-contextual programs $\yvec$,
except for Parse-then-Resolve which uses $\tilde{\yvec}$ instead.
The test input placeholder follows the same form as demonstration input placeholders.
New CFG rules are derived from the program annotations of contextual utterances, \ie $\tilde{\yvec}$ and $\yvec$,
yielding two new grammars $\mathcal{G}_\alpha$ and $\mathcal{G}_\beta$, respectively.
During constrained decoding, 
the joint grammar $\mathcal{G}_1 \cup \mathcal{G}_\alpha$ is used for Parse-then-Resolve,
whereas $\mathcal{G}_1 \cup \mathcal{G}_\beta$ is used for the other three paradigms.
In other words, the adaptation only changes the set of demonstration examples used during prompt instantiation
and augments the CFG used during constrained decoding.

For Rewrite-then-Parse, we can re-use the same grammar $\mathcal{G}_1$ and parser $\mathcal{P}_1$ 
used for non-contextual utterances,
without any annotated programs for contextual utterances.

\subsection{Adaptation via FT}

For FT, the parser $\mathcal{P}_1$ for non-contextual utterances uses an LLM $\mathcal{M}_1$ 
fine-tuned from T5-base 220M \cite{raffel2020exploring}.
To adapt this parser for contextual utterances, 
we continue fine-tuning $\mathcal{M}_1$ on annotated contextual utterances,
except for Rewrite-then-Parse which uses $\mathcal{P}_1$ itself. 
Similar to ICL, different forms of token sequences are used for different paradigms,
\ie
$\hvec\ \lvert\ \uvec\ \lvert\ \yvec$ for Parse-with-Utterance-History,
$\rvec\ \lvert\ \uvec\ \lvert\ \yvec$ for Parse-with-Utterance-History,
and $\uvec\ \lvert\ \tilde{\yvec}$ for Parse-then-Resolve.
The new grammar is constructed identically to ICL as well.

\subsection{Data Annotation Effort}

An important axis when comparing different parsing paradigms is the 
data annotation effort.
For Parse-with-Utterance-History, annotating the non-contextual program for a contextual utterance can be a cognitively demanding task, as it needs to account for the full utterance history.
Data annotation for Parse-with-Reference paradigm is similar to the Parse-with-Utterance-History, though it may be less cognitively intensive because the human annotator only needs to make a a few edits as opposed to performing a full parse.
Compared with Parse-with-Utterance-History, annotations of intermediate programs in the Parse-then-Resolve paradigm are much less context-dependent
and more concise, which potentially makes the parser more data efficient.
However, this comes at a cost of placing a greater burden on the resolver, which uses custom-designed contextual symbols based on the domain;
their expressiveness can greatly affect the 
quality of the annotations and the complexity of the resolver.
Finally, collecting annotations for the the utterance rewriting task is relatively easy and domain independent compared to collecting annotations for parsers which often requires learning a domain-specific language.
\section{Experiments}

\subsection{Data}

Existing CSP datasets are often annotated based on only one or two paradigms,
making it difficult to compare across different paradigms comprehensively.
To address this challenge, we construct a dataset \datasetfullname{} (\dataset{}) derived from a subset of SMCalFlow \cite{SMDataflow2020}.
It contains 31 training and 100 test instances in total.
Each instance consists of a contextual user utterance $\uvec$ during an event-related query (\eg ``\textit{what about Tuesday?}''), 
the corresponding contextual/intermediate program $\tilde{\yvec}$ and non-contextual program $\yvec$,
the utterance history $\hvec$,
the reference program $\rvec$, 
and the rewritten non-contextual utterance $\uvec'$.
The programs ($\yvec$, $\tilde{\yvec}$, $\rvec$) are semi-automatically 
derived from the original SMCalFlow annotations.
The rewritten non-contextual utterances $\uvec'$ are manually annotated by domain experts.
See Appendix~\ref{appendix:dataset} for details of the dataset construction and examples.

We additionally use 8892 training and 100 test instances of non-contextual utterances (\eg ``\textit{do I have any meetings scheduled after Thursday?}''),
each paired with their corresponding non-contextual programs, semi-automatically derived from SMCalFlow as well.
These instances are used to construct and evaluate the base parser $\mathcal{P}_1$ for non-contextual utterances.

\subsection{Experimental Results}

For Parse-with-Reference-Program, we use the oracle reference program, 
which is the non-contextual program of the preceding turn.\footnote{%
It is possible that the reference program is from an earlier turn or does not appear in the history,
though the contextual subset does not contain such examples.}
For Parse-then-Resolve, we assume an oracle resolver is available, which in practice can be implemented as a rule-based system.
The rewriter used for Rewrite-then-Parse is implemented via GPT-3.5, and details are provided in Appendix~\ref{appendix:rewriter}.  
We also consider using the oracle rewritten utterances annotated in the contextual subset of \dataset{}.

\begin{table}[!t]
    \centering
    \begin{tabular}{lll} 
        \toprule 
        {\bf Paradigm}                  & {\bf ICL}         & {\bf FT}\\
        \midrule
        
        Parse-with-Utterance-History    & 51.8              & 81.2 \\
        Parse-with-Reference-Program    & 86.1$^\star$      & 78.2 \\
        Parse-then-Resolve              & 70.5$^\star$      & 82.4 \\ 
        Rewrite-then-Parse              & 65.3$^\star$      & 75.2 \\
        Rewrite-then-Parse (oracle)     & 76.2$^\star$      & 94.0$^\star$ \\
        
        \bottomrule
    \end{tabular}
    \caption{Exact match accuracy on \dataset{} test set. 
    For both ICL and FT, we test each paradigm against the corresponding Parse-with-Utterance-History predictions
    using McNemar's test and show statistically significant ($p < 0.05$) results with $^\star$.
    }
    \label{tab:main_results}
\end{table}

We evaluate the program exact match accuracy on the \dataset{} test set for all paradigms.
\autoref{tab:main_results} presents the experimental results. 
Across all paradigms, FT achieves higher exact match than ICL by 7.9\% to 29.4\% absolute gain. 
For FT, Rewrite-then-Parse with oracle rewritten utterances performs the best.
There is no significant difference among other approaches, 
including Rewrite-then-Parse using the GPT-3.5 rewriter which does not require additional fine-tuning.
For ICL, Parse-with-Reference-Program performs the best, suggesting it is easier for GPT-3.5 to softly edit a program
than parsing directly from natural language. 
Rewrite-then-Parse using oracle rewritten utterances is still better than the remaining approaches. 
By comparing the results of Rewrite-then-Parse, it is clear that improving the rewriter can lead to a corresponding improvement
in parsing accuracy.

We manually examine incorrect predictions made by parsers for contextual utterances 
and identify common error categories:
incorrect top-level program types, alternative parses for the input,
extra constraints, missing constraints, and constraints with incorrect arguments/functions (see Table~\ref{tab:error} for examples).

For ICL, the most common error type is incorrect function calls.
30\% of the errors made by Parse-with-Reference-Program are due to incorrect function use.
In particular, the model struggles with predicting rare functions such as negations, 
potentially because the only knowledge of the target language is from the contextual subset of \dataset{}.

For FT, 
33\% of the errors in Parse-then-Resolve are from incorrect top-level program types. 
Introducing new symbols increases the program space, especially different intermediate programs that have similar functions, suggesting that the design of these specialized contextual symbols is crucial. 
For Parse-with-Utterance-History, we find that 40\% of the errors come from missing constraints,
indicating that jointly learn parsing and consolidating constraints from multiple turns is challenging for the parsing model.
For Rewrite-then-Parse, 55\% of the errors are due to incorrect arguments, 
and 45\% are due to differences in capitalization (\eg the rewriter converts a lowercase name to uppercase) which is arguably less critical.

\begin{table}[!t]
    \centering
    \begin{tabular}{lll} 
        \toprule 
        {\bf Paradigm} 
        & {\bf ICL} & {\bf FT} \\
        \midrule
        
        Parse-with-Utterance-History     & 63.5             & 84.5 \\
        Parse-with-Reference-Program     & 79.5$^\star$     & 83.0 \\       
        Parse-then-Resolve               & 73.5$^\star$     & 85.5 \\       
        Rewrite-then-Parse               & 69.5$^\star$     & 82.0 \\      
        Rewrite-then-Parse (oracle)      & 75.0$^\star$     & 90.5$^\star$ \\
        
        \bottomrule
    \end{tabular}
    \caption{Exact match accuracy on \dataset{} test set combined with non-contextual utterances. 
    For both ICL and FT, we test each paradigm against the corresponding Parse-with-Utterance-History predictions
    using McNemar's test and show statistically significant ($p < 0.05$) results indicated with $^\star$.}
    \label{tab:joint_results}
\end{table}

We also examine the overall parsing accuracy on the joint test set of contextual and non-contextual utterances. We use a binary classifier which takes the user utterance as input and determines whether to use the parser for non-contextual utterances or the parser for contextual utterances.
The classifier is obtained by fine-tuning the RoBERTa-base \cite{liu-etal-2019-roberta} to on \dataset{} utterances.
The overall classification accuracy is 95.5\%.
The results are summarized in \autoref{tab:joint_results}.
We use exact match accuracy as the evaluation metric, where the prediction is treated as correct  
only when classification and parsing are both correct.
\section{Conclusion}

We study a real-world CSP setting, \ie few-shot adaptation for parsing contextual utterances with LLMs,
and compare four different paradigms using both ICL and FT. 
To facilitate the study, we construct a new dataset, \dataset{} with annotations for all paradigms.
Experiments show that ICL with GPT-3.5 usually underperforms FT with T5-base except for Parse-with-Reference-Program,
suggesting GPT-3.5 is good at editing programs via natural language in these data conditions.
Overall, Rewrite-then-Parse stands out as a promising approach for future development of LLM-based CSP,
as it performs as well as other paradigms but require only a few annotated exampels for the rewriter and no additional program annotation.

\section{Limitations}

Due to the cost of collecting program annotations for all paradigms, the size of the \dataset{} test set is relatively small
and we only study dialogues from SMCalFlow.
While the experiments results are informative under significance test, 
it would be useful for future work to conduct a similar study on larger and diverse datasets.

The LLMs used in this work are pre-trained primarily on English, and the \dataset{} also only contains English utterances.
It would be interesting to study the few-shot adaptation problem on other languages.

\section*{Acknowledgements}
We would like to thank Benjamin Van Durme, Matt Gardner, Adam Pauls, and Jason Wolfe for valuable discussions on this paper.

\bibliography{anthology,custom}

\clearpage

\appendix
\setcounter{table}{0}
\renewcommand{\thetable}{A\arabic{table}}

\setcounter{figure}{0}
\renewcommand{\thefigure}{A\arabic{figure}}

\begin{table*}[!th]
\centering
\small
\begin{tabular}{|p{1.25cm}|p{1.25cm}|p{1.25cm}|p{5.5cm}|p{5.5cm}|}
\hline
\textbf{Utterance} & \textbf{Last Utterance} & \textbf{Oracle Rewritten Utterance} & \textbf{Non-Contextual Program} & \textbf{Contextual Program} \\
\hline

What about later next week? &

Did I have any meetings early next week? & Did I have any meetings later next week? &

\begin{lstlisting}[language=Lisp,linewidth=5.5cm]
(QueryEventResponseIsNonEmpty 
(FindEventWrapperWithDefaults 

(Event.duringDateRangeConstraint_?
(LateDateRange (NextWeekList)))))
\end{lstlisting}

&

 \begin{lstlisting}[language=Lisp,linewidth=5.5cm]
(Execute (ReviseConstraint (DefaultRootLocation) (^(Event) ConstraintTypeIntension)

(Event.duringDateRangeConstraint_? (LateDateRange (NextWeekList)))))
\end{lstlisting}  \\
\hline

Actual I meant the day after tomorrow. & Is there any appointments tomorrow? & Is there any appointments the day after tomorrow? &  \begin{lstlisting}[language=Lisp,linewidth=5.5cm]
(QueryEventResponseIsNonEmpty (FindEventWrapperWithDefaults (Event.onDate_? (adjustByPeriod (Tomorrow) (toDays 1)))))
\end{lstlisting} &

 \begin{lstlisting}[language=Lisp,linewidth=5.5cm]
(Execute (ReviseConstraint (DefaultRootLocation) (^(Event) ConstraintTypeIntension) 
(Event.onDate_? (adjustByPeriod (Tomorrow) (toDays 1)))))
\end{lstlisting}  \\ \hline

What about training? & Is there a vacation scheduled for me? & Is there a training scheduled for me? &  \begin{lstlisting}[language=Lisp,linewidth=5.5cm]
(QueryEventResponseIsNonEmpty (FindEventWrapperWithDefaults (Event.subject_? (?~= \"training\"))))\end{lstlisting} &

 \begin{lstlisting}[language=Lisp,linewidth=5.5cm]
(Execute (ReviseConstraint (DefaultRootLocation) (^(Event) ConstraintTypeIntension) (Event.subject_? (?~= \"training\"))))\end{lstlisting}  \\ \hline

\end{tabular}
\caption{Dataset examples.}
\label{tab:examples}
\end{table*}

\begin{table*}[th]
    \centering
    \begin{tabular}{p{2cm}|p{2cm}}
    \toprule
    {\bf Original} & {\bf Simplified} \\
    
    \midrule
    \begin{lstlisting}[language=Lisp,linewidth=7cm]
(& (^($type) EmptyStructConstraint) ($c))
\end{lstlisting} 
    & \begin{lstlisting}[language=Lisp,linewidth=7cm]
($c)
\end{lstlisting} \\
    
    \midrule
    \begin{lstlisting}[language=Lisp,linewidth=7cm]
(& ($c) (^($type) EmptyStructConstraint))
\end{lstlisting} 
    & \begin{lstlisting}[language=Lisp,linewidth=7cm]
($c)
\end{lstlisting} \\

    \midrule
    \begin{lstlisting}[language=Lisp,linewidth=7cm]
(> (size (QueryResponse.results ($response))), 0L)
\end{lstlisting} 
    & \begin{lstlisting}[language=Lisp,linewidth=7cm]
(QueryEventResponseIsNonEmpty ($response))
\end{lstlisting} \\

    \midrule
    \begin{lstlisting}[language=Lisp,linewidth=7cm]
(AttendeeListHasRecipientConstraint (RecipientWithNameLike (^(Recipient) EmptyStructConstraint) (PersonName.apply $name)))
\end{lstlisting} 
    & \begin{lstlisting}[language=Lisp,linewidth=7cm]
(WithAttendeeNamed ($name))
\end{lstlisting} \\

    \midrule
    \begin{lstlisting}[language=Lisp,linewidth=7cm]
(AttendeeListHasRecipient (Execute (refer (extensionConstraint (RecipientWithNameLike (^(Recipient) EmptyStructConstraint) (PersonName.apply $name))))))
\end{lstlisting} 
    & \begin{lstlisting}[language=Lisp,linewidth=7cm]
(WithAttendeeNamed ($name))
\end{lstlisting} \\

    \midrule
    \begin{lstlisting}[language=Lisp,linewidth=7cm]
(AttendeeListExcludeRecipient (Execute (refer (extensionConstraint (RecipientWithNameLike (^(Recipient) EmptyStructConstraint) (PersonName.apply $name))))))
\end{lstlisting} 
    & \begin{lstlisting}[language=Lisp,linewidth=7cm]
(WithoutAttendeeNamed ($name))
\end{lstlisting} \\

    \bottomrule
    \end{tabular}
    \caption{List of sub-tree transformations for simplifying SMCalFlow programs (part 1).}
    \label{tab:simplification_1}
\end{table*}

\begin{table*}[th]
    \centering
    \begin{tabular}{p{2cm}|p{2cm}}
    \toprule
    {\bf Original} & {\bf Simplified} \\

    \midrule
    \begin{lstlisting}[language=Lisp,linewidth=7cm]
(EventAtTime ($event) ($time))
\end{lstlisting} 
    & \begin{lstlisting}[language=Lisp,linewidth=7cm]
(& ($event) (Event.atTime_? ($time)))
\end{lstlisting} \\

    \midrule
    \begin{lstlisting}[language=Lisp,linewidth=7cm]
(EventDuringRangeTime ($event) ($timeRange))))
\end{lstlisting} 
    & \begin{lstlisting}[language=Lisp,linewidth=7cm]
(& ($event) (Event.duringTimeRangeConstraint_? ($timeRange)))
\end{lstlisting} \\

    \midrule
    \begin{lstlisting}[language=Lisp,linewidth=7cm]
(EventOnDate ($date) ($event))
\end{lstlisting} 
    & \begin{lstlisting}[language=Lisp,linewidth=7cm]
(& ($event) (Event.onDate_? ($date)))
\end{lstlisting} \\

    \midrule
    \begin{lstlisting}[language=Lisp,linewidth=7cm]
(EventDuringDateRange ($event) ($dateRange))))
\end{lstlisting} 
    & \begin{lstlisting}[language=Lisp,linewidth=7cm]
(& ($event) (Event.duringDateRangeConstraint_? ($dateRange)))
\end{lstlisting} \\

    \midrule
    \begin{lstlisting}[language=Lisp,linewidth=7cm]
(EventOnDateTime (DateAtTimeWithDefaults (($date) ($time)) ($event)))
\end{lstlisting} 
    & \begin{lstlisting}[language=Lisp,linewidth=7cm]
(& ($event) (& (Event.onDate_? ($date)) (Event.atTime_? ($time))))
\end{lstlisting} \\

    \midrule
    \begin{lstlisting}[language=Lisp,linewidth=7cm]
(EventOnDateAfterTime (($date) ($event) ($time)))
\end{lstlisting} 
    & \begin{lstlisting}[language=Lisp,linewidth=7cm]
(& ($event) (& (Event.onDate_? ($date)) (Event.afterTime_? ($time))))
\end{lstlisting} \\

    \midrule
    \begin{lstlisting}[language=Lisp,linewidth=7cm]
(EventOnDateBeforeTime (($date) ($event) ($time)))
\end{lstlisting} 
    & \begin{lstlisting}[language=Lisp,linewidth=7cm]
(& ($event) (& (Event.onDate_? ($date)) (Event.beforeTime_? ($time))))
\end{lstlisting} \\

    \midrule
    \begin{lstlisting}[language=Lisp,linewidth=7cm]
(EventOnDateFromTimeToTime (($date) ($event) ($time1) ($time2)))
\end{lstlisting} 
    & \begin{lstlisting}[language=Lisp,linewidth=7cm]
(& ($event) (& (Event.onDate_? ($date)) (Event.betweenTimeAndTime_? ($time1) ($time2))))
\end{lstlisting} \\

    \bottomrule
    \end{tabular}
    \caption{List of sub-tree transformations for simplifying SMCalFlow programs (part 2).}
    \label{tab:simplification_2}
\end{table*}

\begin{table*}[th]
    \centering
    \begin{tabular}{p{2cm}|p{2cm}}
    \toprule
    {\bf Original} & {\bf Simplified} \\

    \midrule
    \begin{lstlisting}[language=Lisp,linewidth=7cm]
(EventAfterDateTime (($event) ($dateTime)))
\end{lstlisting} 
    & \begin{lstlisting}[language=Lisp,linewidth=7cm]
(& ($event) (Event.afterDateTime_? ($dateTime)))
\end{lstlisting} \\

    \midrule
    \begin{lstlisting}[language=Lisp,linewidth=7cm]
(EventBeforeDateTime (($event) ($dateTime)))
\end{lstlisting} 
    & \begin{lstlisting}[language=Lisp,linewidth=7cm]
(& ($event) (Event.beforeDateTime_? ($dateTime)))
\end{lstlisting} \\

    \midrule
    \begin{lstlisting}[language=Lisp,linewidth=7cm]
(EventOnDateWithTimeRange (EventOnDate ($date) ($event)) ($timeRange))
\end{lstlisting} 
    & \begin{lstlisting}[language=Lisp,linewidth=7cm]
(& ($event) (& (Event.onDate_? ($date)) (Event.duringTimeRangeConstraint_? ($timeRange))))
\end{lstlisting} \\

    \midrule
    \begin{lstlisting}[language=Lisp,linewidth=7cm]
(EventOnDateWithTimeRange (EventDuringRange ($event) ($dateRange) ($timeRange)))
\end{lstlisting} 
    & \begin{lstlisting}[language=Lisp,linewidth=7cm]
(& ($event) (& (Event.duringDateRangeConstraint_? ($dateRange)) (Event.duringTimeRangeConstraint_? ($timeRange))))
\end{lstlisting} \\

    \midrule
    \begin{lstlisting}[language=Lisp,linewidth=7cm]
(EventDuringRangeDateTime ($event) ($dateTimeRange))
\end{lstlisting} 
    & \begin{lstlisting}[language=Lisp,linewidth=7cm]
(& ($event) (Event.duringDateTimeRangeConstraint_? ($dateTimeRange)))
\end{lstlisting} \\

    \bottomrule
    \end{tabular}
    \caption{List of sub-tree transformations for simplifying SMCalFlow programs (part 3).}
    \label{tab:simplification_3}
\end{table*}
\section{CFG for Constrained Decoding}
\label{appendix:cfg}

The CFG used for constrained decoding can be automatically derived from function definitions and types used in the domain. 
For example, given a function $\texttt{FN}(\text{arg}_1, \cdots, \text{arg}_N)$ with corresponding argument types $\tau_1, \cdots, \tau_N$ and output type $\tau_O$,
we can automatically derive a CFG rule %
$\textrm{NT}_{\tau_O} \rightarrow 
( \ \texttt{FN} \  (
\ \textrm{NT}_{\tau_1}
\cdots 
\textrm{NT}_{\tau_N}
\ ) \  )$
where $\textrm{NT}_{\tau_i}$ denotes the non-terminal symbol for the type $\tau_i$,
and the function name \texttt{FN} and the parentheses are terminal symbols 
in $\mathcal{G}$.
For each primitive type (\eg ``string'', ``number''), we additionally define
CFG rules to expand the non-terminal of the primitive type to terminals representing acceptable values of the type
(sometimes using regular expressions).

\section{Dataset Construction and Examples}
\label{appendix:dataset}

The original SMCalFlow data only provide annotations of \textit{contextual programs} for individual utterances.
We develop a heuristic-based implementation of \texttt{NewClobber} and \texttt{ReviseConstraint} to propose candidates of 
the corresponding non-contextual programs.
Specifically, given the non-contextual program
\begin{lstlisting}[language=Lisp]
(NewClobber (
    (intension) 
    (slotConstraint) 
    (value)))
\end{lstlisting} 
we modify the non-contextual program of the preceding turn by replacing its fragment satisfying 
the \texttt{slotConstraint} with the new fragment \texttt{value}.
Similarly, given the non-contextual program
\begin{lstlisting}[language=Lisp]
(ReviseConstraint (
  (rootLocation) 
  (oldLocation) 
  (newConstraint)))
\end{lstlisting} 
we modify the non-contextual program of the preceding turn by replacing a fragment \texttt{oldConstraint}
which satisfies the \texttt{oldLocation} and is governed by a bigger fragment satisfying the \texttt{rootLocation} 
with a new fragment
\begin{lstlisting}[language=Lisp]
(& ((oldConstraint) (newConstraint)))
\end{lstlisting} 
\ie conjoining the two constraints regardless whether they conflicts with each other.
For both cases, when there are multiple possible replacements, all resulting candidates are proposed.
These candidates are manually reviewed and edited by the authors to finalize non-contextual program annotations.
For example, if \texttt{newConstraint} contradicts with a part of \texttt{oldConstraint}, 
we drop the such conflicting parts in the \texttt{oldConstraint}.

Furthermore, as noted by \citet{meron-2022-simplifying}, the original annotations of SMCalFlow can be complex
and contain many boilerplate segments.
Therefore, we use heuristics to simplify the original annotations to obtain programs that are shorter and potentially easier to read and predict.
Similar to \citet{meron-2022-simplifying}, the simplification was implemented via a set of tree transformation rules, which convert
specific sub-trees of the original program into simplified sub-trees.
The list of sub-tree transformations are provided in \autoref{tab:simplification_1}--\autoref{tab:simplification_3}.

Two data specialists are asked to produce the annotations for the rewritten non-contextual utterances in the contextual subset.
They are provided with instructions and training materials,
which explains how to rewrite a contextual user utterance with its preceding utterance into a single non-contextual utterance.
Each example takes 10 to 30 seconds to annotate. Additionally, annotators were asked to provide a confidence from 0 (least confident) to 3 (most confident) in the rewritten utterance. The average confidence was 2.9. Then they are asked to review the each other's annotations and answer whether they agree with each other. In our pilot data collection, the agreement rate between the two data specialists was 93.3\%.

Table~\ref{tab:examples} provides some examples from in \dataset{}.

\section{Fine-tuning Experiment Hyperparameters}
\label{appendix:hyperparameters}

For fine-tuning, we employ the Adafactor optimizer \cite{shazeer2018adafactor} and set the batch size to 32.
The slanted triangular learning rate scheduler \cite{howard-ruder-2018-universal} is used
with a maximum learning rate of $10^{-5}$ and 1000 warmup steps.
We fine-tune $\mathcal{M}_0$ for 10000 steps on the non-contextual subset to obtain $\mathcal{M}_1$, 
and another 10000 steps on the corresponding data to obtain the models for individual paradigms.
For constrained decoding, the maximum output sequence length is 1000.

\section{Rewriter Implementation}
\label{appendix:rewriter}

The rewriter used for Rewrite-then-Parse is implemented via GPT-3.5 (\texttt{text-davinci-003}).  
The prompt template is shown below, where placeholders 
\{H1\}, \{H2\}, \ldots\ are for the utterance history (\ie the preceding utterances),
\{X1\}, \{X2\}, \ldots\ are for contextual user utterances,
\{Z1\}, \{Z2\}, \ldots\ are for rewritten non-contextual utterances,
and \{H'\} and \{X'\} are for test input.

\begin{lstlisting}[breaklines=true]
Combine the utterances into a single utterance
with the meaning of the last utterance.


Last Utterance: {H1}
Current Utterance: {X1}
Rewritten Utterance: {Z1}

Last Utterance: {H2}
Current Utterance: {X2}
Rewritten Utterance: {Z2}

...

Last Utterance: {H'}
Current Utterance: {X'}
Rewritten Utterance:
\end{lstlisting}

We sample 8 demonstration examples are sampled uniformly from the contextual subset training instances.
Greedy decoding is used with 50 maximum tokens and no frequency or presence penalty. The BLEU score using the oracle rewritten utterances as reference is 93.6.

\begin{table*}[!th]
\centering
\small
\begin{tabular}{|p{1.5cm}|p{2cm}|p{2cm}|}
\hline
\textbf{Error Type} & \textbf{Gold} & \textbf{Predicted} \\
\hline

Top-level Incorrect & \begin{lstlisting}[language=Lisp,linewidth=7cm]
(Execute (ReviseConstraint (DefaultRootLocation) (^(Event) ConstraintTypeIntension) (& (Event.attendees_? (WithAttendeeNamed "kim")) (& (Event.onDate_? (Tomorrow)) (Event.subject_? (?~= "lunch meeting"))))))
\end{lstlisting} & \begin{lstlisting}[language=Lisp,linewidth=7cm]
(Execute (NewClobber (DefaultIntension) (^(Recipient) ConstraintTypeIntension) (intension (RecipientWithNameLike (^(Recipient) EmptyStructConstraint) (PersonName.apply "kim")))))
\end{lstlisting} \\
\hline
Alternate parse & \begin{lstlisting}[language=Lisp,linewidth=7cm]
(FindEventWrapperWithDefaults (& (Event.attendees_? (WithAttendeeNamed "Barry")) (Event.start_? (DateTime.date_? (?= (Tomorrow))))))

\end{lstlisting} & \begin{lstlisting}[language=Lisp,linewidth=7cm]
(FindEventWrapperWithDefaults (& (Event.attendees_? (WithAttendeeNamed "Barry")) (Event.onDate_? (Tomorrow))))
\end{lstlisting} \\
\hline

Extra Constraint & \begin{lstlisting}[language=Lisp,linewidth=7cm]
(QueryEventResponseIsNonEmpty (FindEventWrapperWithDefaults (Event.attendees_? (& (WithAttendeeNamed "Marco") (WithAttendeeNamed "Peyton")))))

\end{lstlisting} & \begin{lstlisting}[language=Lisp,linewidth=7cm]
(QueryEventResponseIsNonEmpty (FindEventWrapperWithDefaults (& (Event.attendees_? (WithAttendeeNamed "Peyton")) (& (Event.attendees_? (WithAttendeeNamed "Marco")) (Event.duringDateRangeConstraint_? (FullMonthofMonth (Date.month (Today))))))))
\end{lstlisting} \\
\hline

Missing \\Constraint & \begin{lstlisting}[language=Lisp,linewidth=7cm]
(QueryEventResponseIsNonEmpty (FindEventWrapperWithDefaults (& (Event.attendees_? (WithAttendeeNamed "Bob")) (& (Event.duringTimeRangeConstraint_? (Afternoon)) (Event.onDate_? (Tomorrow))))))

\end{lstlisting} & \begin{lstlisting}[language=Lisp,linewidth=7cm]
(QueryEventResponseIsNonEmpty (FindEventWrapperWithDefaults (& (Event.attendees_? (WithAttendeeNamed "Bob")) (Event.duringTimeRangeConstraint_? (Afternoon)))))\end{lstlisting} \\

\hline

Constraint With Incorrect Function & \begin{lstlisting}[language=Lisp,linewidth=7cm]
(Execute (NewClobber (DefaultIntension) (extensionConstraint (^(LocationKeyphrase) AlwaysTrueConstraint)) (intension (LocationKeyphrase.apply "EVO"))))\end{lstlisting} & \begin{lstlisting}[language=Lisp,linewidth=7cm]
(Execute (NewClobber (DefaultIntension) (^(Recipient) ConstraintTypeIntension) (intension (RecipientWithNameLike (^(Recipient) EmptyStructConstraint) (PersonName.apply "EVO")))))\end{lstlisting} \\
\hline

Constraint With Incorrect Argument  & \begin{lstlisting}[language=Lisp,linewidth=7cm]
(QueryEventResponseIsNonEmpty (FindEventWrapperWithDefaults (Event.onDate_? (adjustByPeriod (Tomorrow) (toDays 1)))))
\end{lstlisting} & \begin{lstlisting}[language=Lisp,linewidth=7cm]
(QueryEventResponseIsNonEmpty (FindEventWrapperWithDefaults (Event.onDate_? (adjustByPeriod (Tomorrow) (toDays 2)))))\end{lstlisting} \\
\hline

\end{tabular}
\caption{Error examples with gold parses.}
\label{tab:error}
\end{table*}

\end{document}